%% file: PSviaDL.tex
\documentclass[9pt]{article}
\usepackage{spconf}

\usepackage{amsmath}
\usepackage{amssymb}
\usepackage{graphicx}
\usepackage{multirow}
\usepackage[T1]{fontenc}
\usepackage{url}
\usepackage{cite}
\usepackage{mathtools}
\usepackage{mdwmath}
\usepackage{mdwtab}
\usepackage{comment}
\usepackage{enumerate}
\usepackage{scrextend}
\usepackage{caption}
\usepackage{subcaption}
\usepackage{color}
\usepackage{soul}

\DeclareMathOperator*{\argmin}{arg\,min}

\title{ROBUST PHOTOMETRIC STEREO USING LEARNED IMAGE AND GRADIENT DICTIONARIES}
\name{Andrew J. Wagenmaker, Brian E. Moore, and Raj Rao Nadakuditi
\thanks{This work was supported in part by the following grants: ONR grant N00014-15-1-2141, DARPA Young Faculty Award D14AP00086, and ARO MURI grants W911NF-11-1-0391 and 2015-05174-05.}}
\address{Department of EECS, University of Michigan, Ann Arbor, MI, USA}

\begin{document}

\maketitle

\begin{abstract}
Photometric stereo is a method for estimating the normal vectors of an object from images of the object under varying lighting conditions. Motivated by several recent works that extend photometric stereo to more general objects and lighting conditions, we study a new robust approach to photometric stereo that utilizes dictionary learning. Specifically, we propose and analyze two approaches to adaptive dictionary regularization for the photometric stereo problem. First, we propose an image preprocessing step that utilizes an adaptive dictionary learning model to remove noise and other non-idealities from the image dataset before estimating the normal vectors. We also propose an alternative model where we directly apply the adaptive dictionary regularization to the normal vectors themselves during estimation. We study the practical performance of both methods through extensive simulations, which demonstrate the state-of-the-art performance of both methods in the presence of noise.
\end{abstract}


\section{INTRODUCTION} \label{sec:intro}
\input{sec_intro.tex}

\section{PHOTOMETRIC STEREO PROBLEM} \label{sec:prelims}
\input{sec_prelim.tex}

\section{DICTIONARY LEARNING MODELS} \label{sec:dl}
\input{sec_dl.tex}

\section{RESULTS} \label{sec:examples}
\input{sec_ex.tex}

\vspace{-2mm}
\section{CONCLUSION} \label{sec:conclusion}
\vspace{-1mm}
In this work, we investigated two dictionary learning based methods for robust photometric stereo. Each method seeks to represent some form of our data---either the original images or the estimated normal vectors---as sparse with respect to an adaptive (learned) dictionary. We showed that both approaches are significantly more robust to noise than existing methods. The results presented here indicate that DLPI usually outperforms DLNV, but which method performs best in general may depend on underlying properties of the data. We leave this nuanced investigation for future work.

\bibliographystyle{IEEEbib}
\ninept
\bibliography{PSviaDL}

\end{document}

%% file: sec_intro.tex

Photometric stereo \cite{woodham1980} is a method for estimating the normal vectors of an object from images of the object under varying lighting conditions. Since its inception, a significant amount of work has been done extending photometric stereo to more general conditions. This body of work has been divided into two primary areas. Uncalibrated photometric stereo seeks to solve the photometric stereo problem when the lighting directions are unknown \cite{hayakawa1994,belhumeur1999,yuille1999,georghiades2003}, while robust photometric stereo algorithms attempt to estimate the normal vectors of an object when the surface violates the assumptions of the underlying model (usually, the Lambertian reflectance model). In this work, we are primarily concerned with the latter problem.

The Lambertian reflectance model states that the observed intensity of a point on a surface is linearly proportional to the direction the surface is illuminated and the object's normal vectors \cite{woodham1980}. While this assumption holds in some cases, shadows, specularities, and other non-idealities can cause this model to break down. A variety of techniques have been developed to compensate for these corruptions. Several works seek to model non-Lambertian effects as outliers and employ a framework to estimate these outliers and discard them from the data, leaving only Lambertian data behind \cite{barsky2003,chandraker2007,verbiest2008,yu2010,wu2010}. Of particular interest in this category are the more recent works by Wu \textit{et al}. \cite{wu2011} and Ikehata \textit{et al}. \cite{ikehata2012}. These works model the outliers as a sparse matrix, casting the problem as a matrix completion problem, and use robust PCA or sparse regression, respectively, to solve for some Lambertian representation of the data. Other works have proposed more complicated reflectance models to account for non-Lambertian effects, eliminating the need to discard non-ideal data \cite{oren1995,hertzmann2005,alldrin2007_2,chung2008,alldrin2008,seitz2010,higo2010,shi2012,chandraker2013}.  The current state-of-the-art in this category are the works by Ikehata \textit{et al}. \cite{ikehata2014} and Shi \textit{et al}. \cite{shi2014}.

In this work, we propose two new approaches for photometric stereo that are robust to noisy data. Our methods utilize a dictionary learning model \cite{elad2006image,aharon2006rm} to handle non-idealities and impose some adaptive structure on the data. Our approach is motivated in part by the recent success of dictionary learning in other imaging domains \cite{ravishankar2011mr,ravishankar2016lassi}.
We demonstrate the viability and performance of our methods on several datasets with varying degrees of non-ideality---including the recently proposed DiLiGenT dataset \cite{shi2016}---comparing them to the performance of state-of-the-art methods. In particular, we investigate the ability of our methods to handle general, non-sparse errors and noise.

The rest of this paper is organized as follows. Section 2 introduces the photometric stereo problem. Section 3 presents our dictionary learning based formulations and details their implementation. Finally, in Section 4, we demonstrate the performance of our methods on a variety of datasets.

%% file: sec_prelim.tex

The Lambertian reflectance model states that, given an image of a \emph{Lambertian} object, the light intensity observed at point $(x,y)$ on the surface satisfies
\begin{equation} \label{ps1}
I(x,y) = \rho(x,y) \ell^T n(x,y),
\end{equation}
where $I(x,y)$ is the image intensity, $\ell \in \mathbb{R}^3$ is the direction of the light source incident on the surface, $n(x,y)$ is the normal vector of the surface, and $\rho (x,y)$ is the surface albedo---a measure of the reflectivity of the surface. 

If we fix the position of a camera facing our surface and vary the position of the light source over $d$ unique locations, we can write $d$ equations of the form \eqref{ps1}, which can be stacked to form the system of equations
\begin{equation} \label{ps2}
[I_1(x,y)~\ldots~I_d(x,y)]^T = [\ell_1~\ldots~\ell_d]^T \rho(x,y) n(x,y).
\end{equation}
Assuming each of the $d$ images has size $m_1 \times m_2$, Equation \eqref{ps2} can then be solved $m_1 m_2$ times to obtain the normal vectors of the object at each point on its surface. These equations can also be combined into a single matrix equation. Indeed, let us define the observation matrix
\begin{equation} \label{ddef}
Y := \left [\textbf{vec}(I_1)~\ldots~\textbf{vec}(I_d) \right ] \in \mathbb{R}^{m_1 m_2 \times d},
\end{equation}
where $\textbf{vec}(I_j) := [I_j(1,1)~\ldots~I_j(m_1,m_2)]^T$. Assuming our light source is at infinity and there is no variation in illumination from point to point on our object, one can succinctly express \eqref{ps2} as
\begin{equation} \label{ps3}
Y = NL,
\end{equation}
where $N = [\rho(1,1) n(1,1)~\ldots~\rho(m_1,m_2) n(m_1,m_2)]^T \in \mathbb{R}^{m_1 m_2 \times 3}$ and $L = [\ell_1~\ldots~\ell_d] \in \mathbb{R}^{3 \times d}$. For simplicity, we assume that $\|\ell_k\|_2=1$, and, without loss of generality, we assume that $n(x,y)$ are \emph{unit} normals.

Given $d \geq 3$ images and their corresponding light directions, one can solve \eqref{ps3} exactly to obtain the normal vector matrix $N$, from which one can compute the full 3D representation of the underlying surface \cite{simchony1990}.

In theory, \eqref{ps3} should hold exactly for a Lambertian surface, but, in practice, due to noise and other non-idealities, one only expects that $Y \approx NL$. In the latter case, one can instead collect $d > 3$ measurements and solve the overdetermined least squares problem
\begin{equation} \label{eq:ls}
\min_{N} \ \left \| Y - NL \right \|_F^2,
\end{equation}
which has the convenient closed-form solution $\hat{N} = YL^\dagger$, where $\dagger$ denotes the Moore-Penrose pseudoinverse.

%% file: sec_dl.tex

In this section, we propose two adaptive dictionary learning methods for estimating the normal vectors of a surface from (possibly) noisy images, $Y$. Intuitively, these models seek to learn a locally sparse representation of the data with respect to a collection of learned basis ``atoms'' that capture the underlying local structure of the data.

\subsection{Preprocessing Images through Dictionary Learning (DLPI)}
Our first approach applies dictionary learning to the data in a preprocessing step before estimating the normal vectors. This formulation represents the input image data $Y$ as locally sparse in an adaptive dictionary domain---thereby removing non-idealities that are not well-represented by the dictionary. Specifically, we propose to solve the problem
\begin{align} \label{eq:dlpi}
\min_{v,B,D} & ~ \frac{1}{2} \left \| y -  v \right \|_2^2 + \lambda \left(\textstyle\sum_{j=1}^c \left \| P_j v  - D b_j \right \|_2^2 +  \mu^2 \left \| B \right \|_0 \right) \nonumber \\
\text{s.t.} & ~ \left \| d_i \right \|_2 = 1, \ \ \left \| b_j \right \|_{\infty} \leq a, ~~ \forall i,j.
\end{align}
Here, $y = \textbf{vec}(Y)$ and $P_j$ is a matrix that extracts a (vectorized) 3D patch of dimensions $c_x \times c_y \times c_z$ from $v$, where $c_x$ and $c_y$ are the dimensions of the patches extracted from each image and $c_z$ is the number of distinct images whose patches are combined to form the 3D patch. $D \in \mathbb{R}^{c_x c_y c_z \times K}$ is a dictionary matrix whose columns $d_i$ are the (learned) dictionary atoms, and $B \in \mathbb{R}^{K \times c}$ is a sparse coding matrix whose columns $b_j$ define (usually sparse) linear combinations of dictionary atoms used to represent each patch. Also, $\left \|\cdot\right \|_0$ is the familiar $\ell_0$ ``norm", and $\lambda,\mu > 0$ are parameters. 

We impose the constraint $\|b_j\|_{\infty} \leq a$, where $a$ is typically very large, since \eqref{eq:dlpi} is non-coercive with respect to $B$, but the constraint is typically inactive in practice \cite{sairajfes2}. Without loss of generality, we impose a unit-norm constraint on the dictionary atoms $d_i$ to avoid scaling ambiguity between $D$ and $B$ \cite{kar}.
We allow the possibility that patches from $c_z > 1$ input images can be combined into a 3D patch to allow the dictionary atoms to learn correlated features between images, but one can set $c_z = 1$ to work with 2D per-image patches.

Once we have solved \eqref{eq:dlpi}, we reshape $v$ (back) into an $m_1 m_2 \times d$ matrix whose columns are vectorized (now preprocessed) images, and then we estimate the associated normal vectors using the standard least squares model \eqref{eq:ls}. Henceforth, we refer to this approach as the Dictionary Learning with Preprocessed Imgaes (DLPI) method.

\subsection{Normal Vector Computation through Dictionary Learning (DLNV)}
We next propose modifying \eqref{eq:ls} by applying an adaptive dictionary regularization term to the normal vectors, $N$, under the Lambertian model \eqref{eq:ls}. Specifically, we propose to solve the problem
\begin{align} \label{eq:dlnv}
\min_{n,B,D} & ~ \frac{1}{2} \left \| y - A n \right \|_2^2 + \lambda \left ( \begin{matrix} \sum_{j=1}^w \left \| P_j n - D b_j \right \|_2^2 \end{matrix} + \mu^2 \left \| B \right \|_0 \right ) \nonumber \\
\text{s.t.} & ~ \left \| d_i \right \|_2 = 1, \ \ \left \| b_j \right \|_{\infty} \leq a, ~~ \forall i,j.
\end{align}
Here $y = \textbf{vec}(Y)$, $A = L^T \otimes I$---where $\otimes$ denotes the Kronecker product and $I$ is the $m_1 m_2 \times m_1 m_2$ identity matrix---and $n = \textbf{vec}(N)$. Also, $P_j$ denotes a patch extraction matrix that extracts (vectorized) patches of dimensions $w_x \times w_y \times w_z$ from $n$. All other terms are defined analogously to the corresponding terms in \eqref{eq:dlpi} with appropriate dimensions. 

The dictionary learning terms in \eqref{eq:dlnv} encourage the estimated normal vectors to be well-represented by sparse linear combinations of a few (learned) dictionary atoms. Intuitively,  this acts as an adaptive regularization that yields normal vectors that are more robust to noise and other non-idealities in the data.
Henceforth, we refer to this approach as the Dictionary Learning on Normal Vectors (DLNV) method.

\subsection{Algorithms for DLPI and DLNV} \label{sec:dl_sol}
We propose solving \eqref{eq:dlpi} and \eqref{eq:dlnv}, respectively, via block coordinate descent-type algorithms where we alternate between updating $n$ and $v$, respectively, with $(D,B)$ fixed and then updating $(D,B)$ with $n$ or $v$ held fixed. We omit the $(D,B)$ updates here due to space considerations, but the precise update expressions can be found in \cite{sairajfes2,dinokat2016}. 

\vspace{-2mm}
\subsubsection{$v$ update}
Solving \eqref{eq:dlpi} for $v$ with $D$ and $B$ fixed yields the problem
\begin{equation} \label{vupdate}
\min_{v} ~ \frac{1}{2} \left \| y -  v \right \|_2^2 + \begin{matrix} \lambda  \sum_{j=1}^c \left \| P_j v  - D b_j \right \|_2^2 \end{matrix}.
\end{equation}
Equation \eqref{vupdate} is a simple least squares problem whose solution $v$ satisfies the normal equation
\begin{equation}\label{eq:vnormal}
\left ( I + \begin{matrix} 2 \lambda \sum_{j=1}^c P_j^T P_j \end{matrix} \right ) v = y + \begin{matrix} 2 \lambda \sum_{j=1}^c P_j^T D b_j \end{matrix},
\end{equation}
where $I$ denotes the identity matrix. The matrix pre-multiplying $v$ in \eqref{eq:vnormal} is diagonal, so its inverse can be cheaply computed, allowing us to efficiently update $v$.

\vspace{-2mm}
\subsubsection{$n$ update}
On the other hand, solving \eqref{eq:dlnv} for $n$ with $D$ and $B$ fixed yields the problem
\begin{equation}\label{nupdate}
\min_{n} ~ \frac{1}{2} \left \| y - A n \right \|_2^2 + \begin{matrix} \lambda \sum_{j=1}^w  \left \| P_j n - D b_j \right \|_2^2 \end{matrix}.
\end{equation}
Note that while \eqref{nupdate} is also a least squares problem, its normal equation cannot be easily inverted as in \eqref{vupdate} due to the presence of the $A$ matrix. We therefore adopt a proximal gradient scheme \cite{parboyd}. The cost function in \eqref{nupdate} can be written in the form $f(n)+g(n)$ where $f(n) = \frac{1}{2} \left \| y - A n \right \|_2^2$ and $g(n) = \lambda \sum_{j=1}^w \left \| P_j n - D b_j \right \|_2^2$. The proximal updates thus become
\begin{equation} \label{nproxupdate}
n^{k+1} = \textbf{prox}_{\tau g}(n^{k} - \tau \nabla f(n^{k})),
\end{equation}
where
\begin{equation} \label{nprox}
\textbf{prox}_{\tau g} (z) := \argmin_{x} \ \frac{1}{2} \left \| z - x \right \|_2^2 + \tau g(x).
\end{equation}
Define $\tilde{n}^{k} := n^{k} - \tau \nabla f(n^{k})$. Then \eqref{nproxupdate} and \eqref{nprox} imply that $n^{k+1}$ satisfies the normal equation
\begin{equation}\label{eq:nnormal}
 \left(I +  \begin{matrix} 2 \tau \lambda \sum_{j=1}^w P_j^T P_j \end{matrix} \right ) n^{k+1} = \tilde{n}^{k} + \begin{matrix} 2 \tau \lambda \sum_{j=1}^w P_j^T D b_j  \end{matrix}.
\vspace{1mm}
\end{equation}
As in \eqref{eq:vnormal}, the matrix multiplying $n^{k+1}$ in \eqref{eq:nnormal} is diagonal and can be efficiently inverted, yielding $n^{k+1}$. Note that proximal gradient is one of a wealth of available iterative schemes for minimzing  the (quadratic) objective \eqref{nupdate}.

%% file: sec_ex.tex

We now empirically demonstrate the performance of our proposed methods on several real-world datasets. To obtain quantitative results, we rely primarily on the DiLiGenT dataset \cite{shi2016}. This dataset contains images of a variety of surfaces and provides the true normal vectors of each object, allowing us to evaluate the performance of each method against a ground truth. We quantify error by measuring the mean angular difference between true normal vectors and estimated normal vectors. 

For each experiment, we compare the results of our method to Wu \textit{et al}.'s robust PCA (RPCA) method \cite{wu2011}, Ikehata \textit{et al}.'s sparse regression (SR) method \cite{ikehata2012}, and Ikehata \textit{et al}.'s constrained bivariate regression (CBR) method \cite{ikehata2014}. We also compare against the simple least squares (LS) method \eqref{eq:ls}. With the exception of LS, each method contains one or more tunable parameters that dictate their performance. For each method, we sweep the parameters across a wide range of values, including any values recommended by the authors in this sweep. The reported results are the errors produced by the optimal parameter values.

To evaluate the ability of our method to robustly reject non-idealities, we add Poisson noise to the images in the original datasets. In each case, we run the experiment over multiple noise realizations and average the results.

\begin{figure*}
\centering
\begin{minipage}{0.46\textwidth}
\includegraphics[width=\columnwidth]{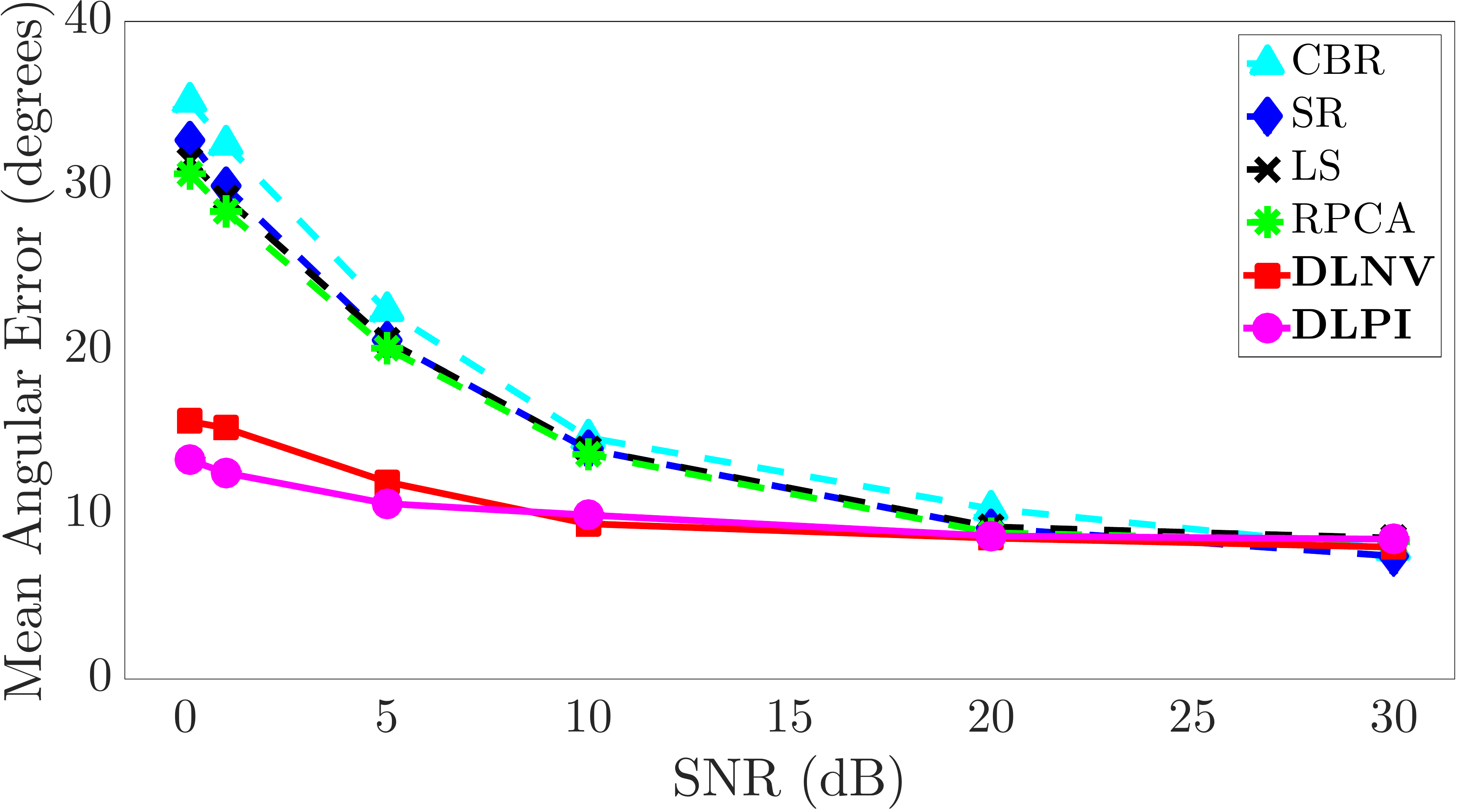}
\caption{Sweeping SNR on the DiLiGenT Cat \cite{shi2016} dataset with 20 images.}
\label{fig:sweep_snr_20_na_0_Dcats_all}
\end{minipage}
\quad 
\begin{minipage}{0.46\textwidth}
\includegraphics[width=\columnwidth]{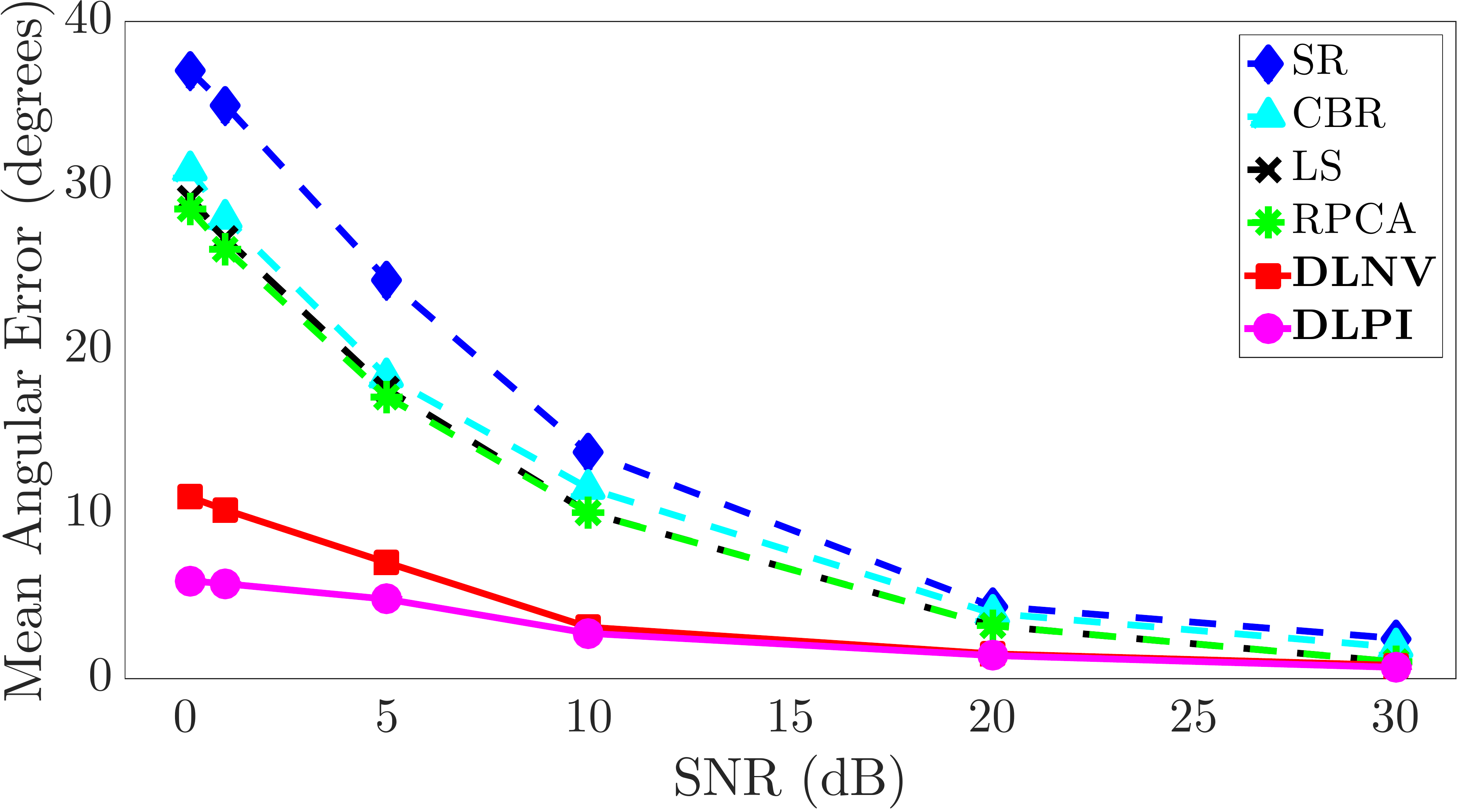}
\caption{Sweeping SNR on the Hippo dataset \cite{xiong2015shading} with 20 images.}
\label{fig:sweep_snr_20_na_0_hippo_all}
\end{minipage}

\vspace{2mm}

\begin{subfigure}[b]{0.105\textwidth}
  \includegraphics[width=\textwidth]{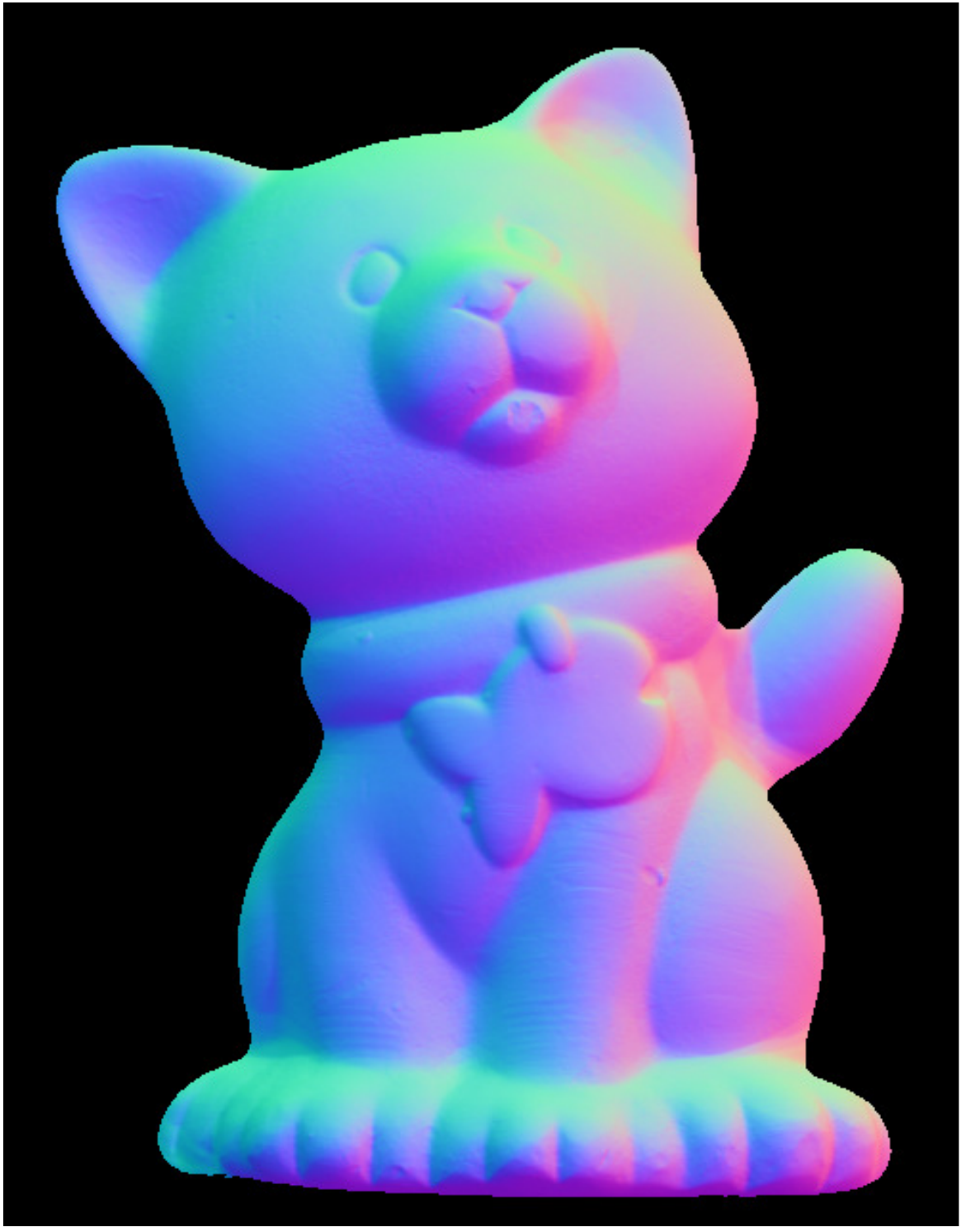}
  \caption{Truth}
\end{subfigure}
\hspace{-2mm}
\begin{subfigure}[b]{0.105\textwidth}
  \includegraphics[width=\textwidth]{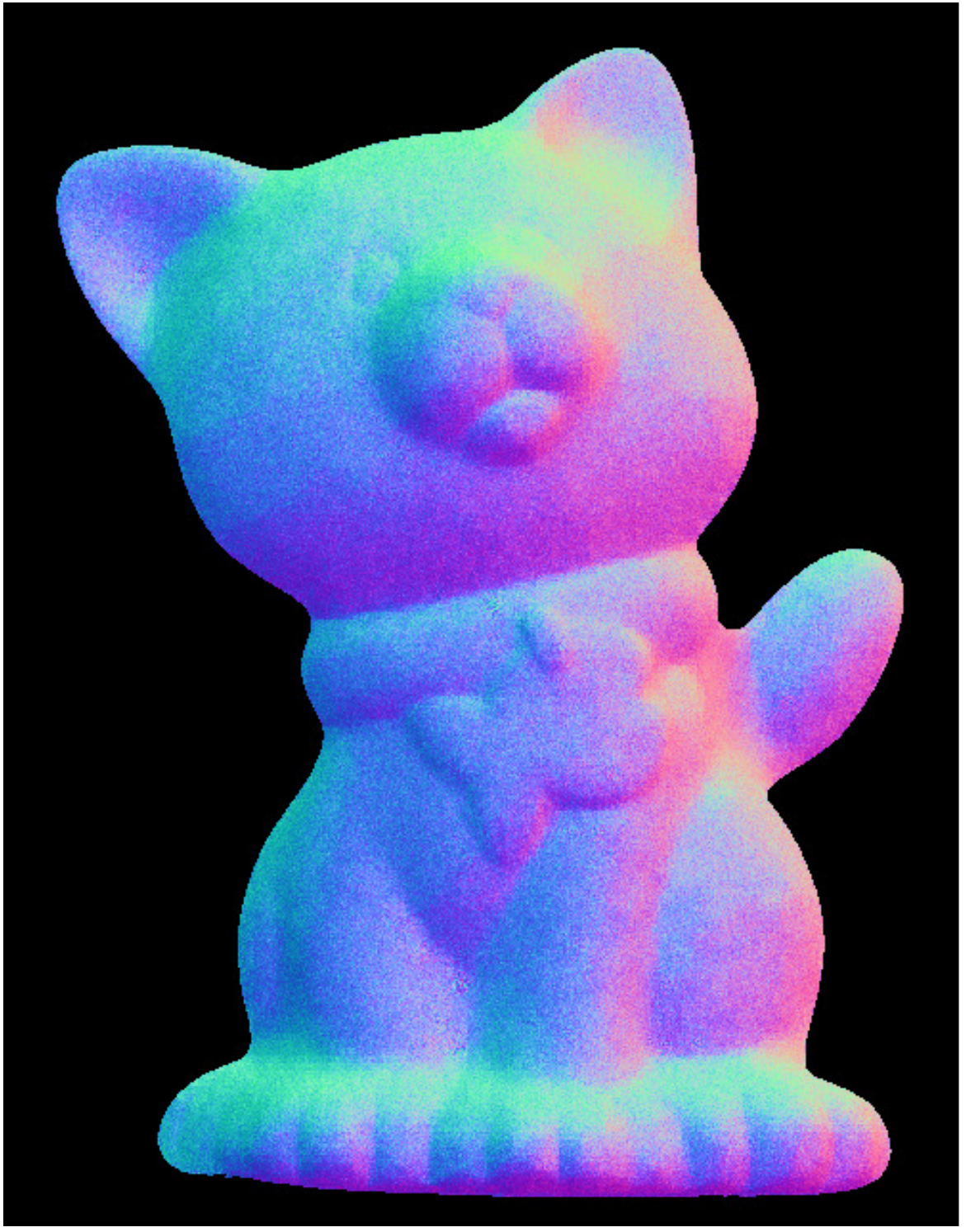}
  \caption{\textbf{DLNV}}
\end{subfigure}
\hspace{-2mm}
\begin{subfigure}[b]{0.105\textwidth}
  \includegraphics[width=\textwidth]{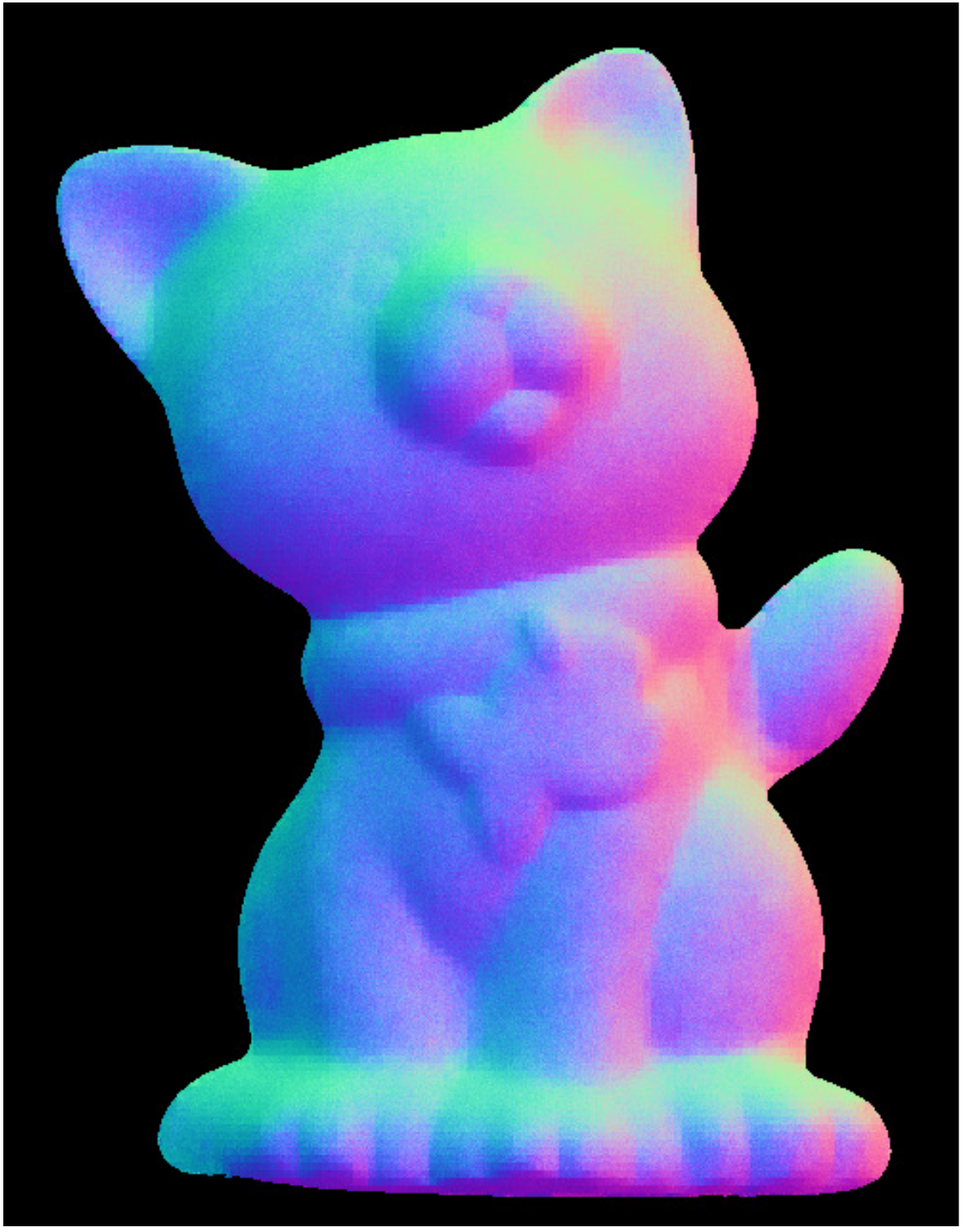}
  \caption{\textbf{DLPI}}
\end{subfigure}
\begin{subfigure}[b]{0.105\textwidth}
  \includegraphics[width=\textwidth]{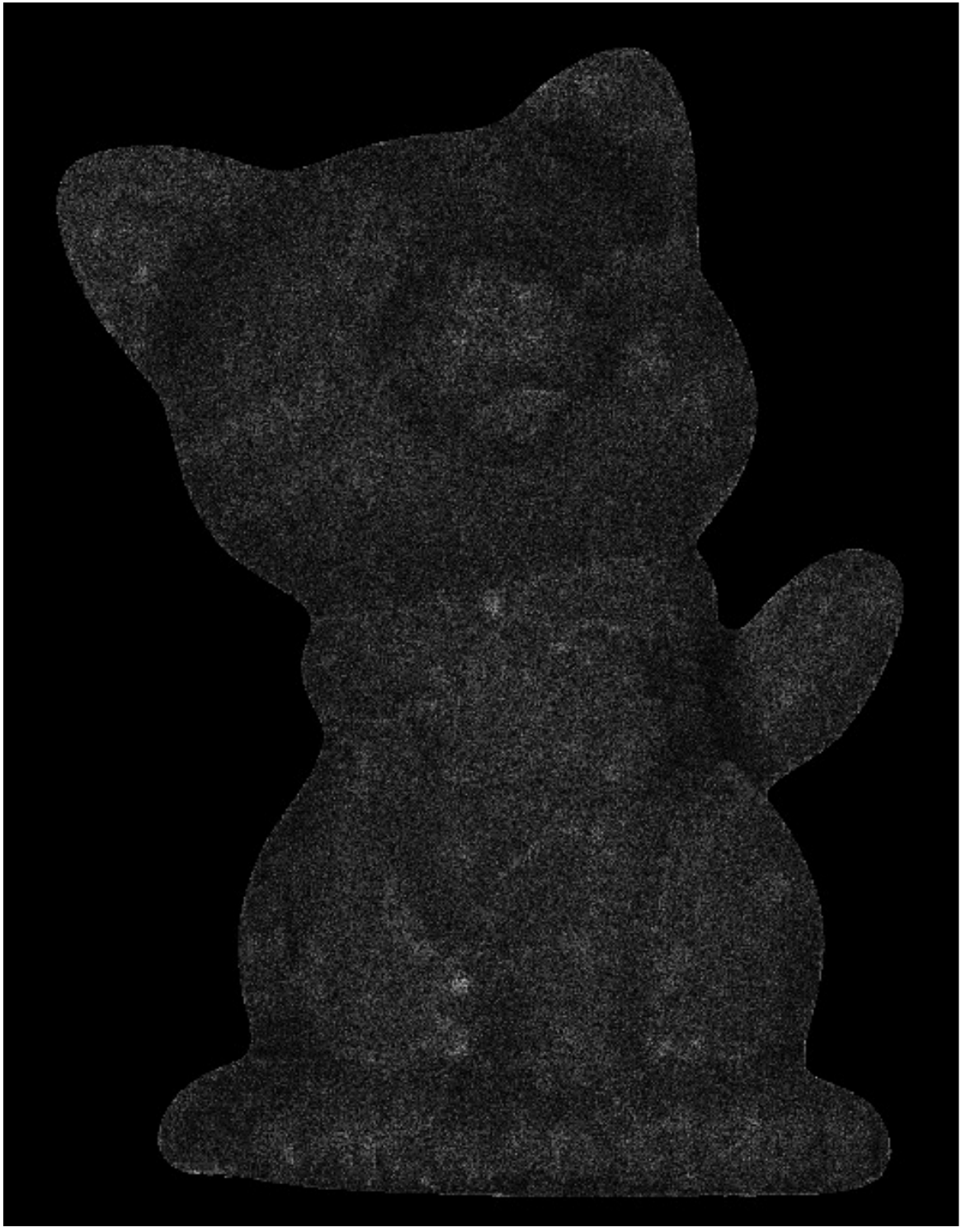}
  \caption{DLNV}
\end{subfigure}
\hspace{-2mm}
\begin{subfigure}[b]{0.105\textwidth}
  \includegraphics[width=\textwidth]{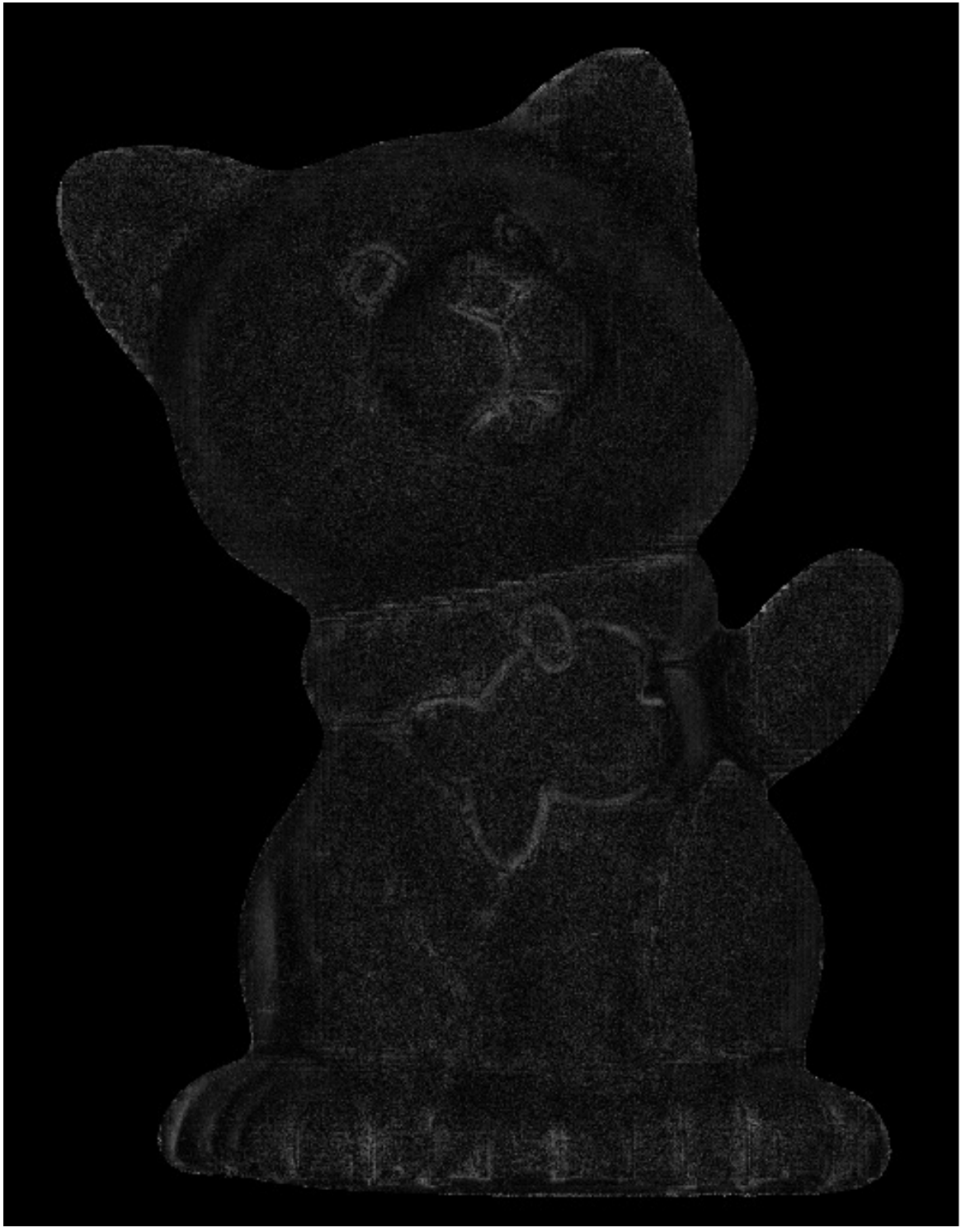}
  \caption{DLPI}
\end{subfigure}
\hspace{-2mm}
\begin{subfigure}[b]{0.105\textwidth}
  \includegraphics[width=\textwidth]{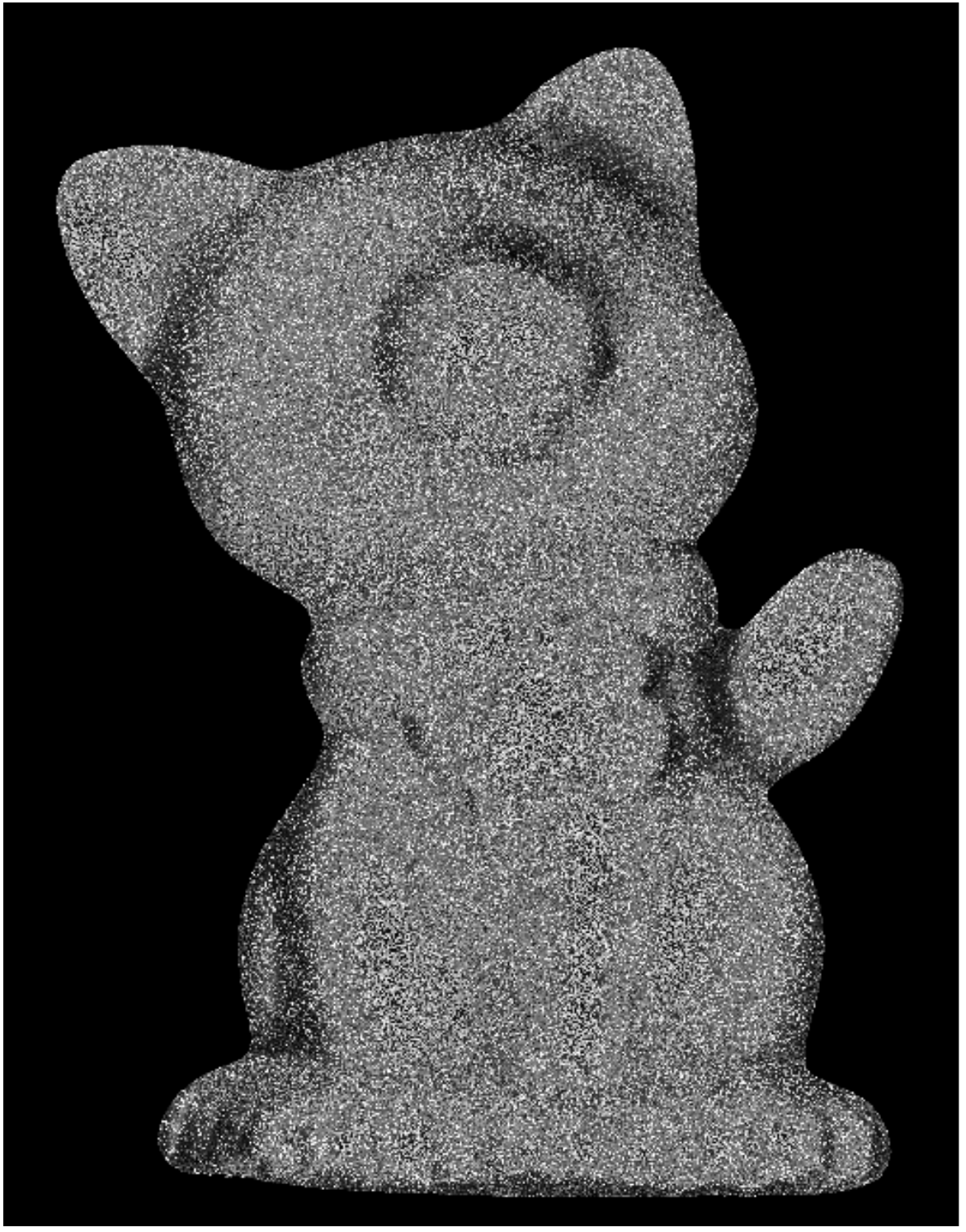}
  \caption{SR}
\end{subfigure}
\hspace{-2mm}
\begin{subfigure}[b]{0.105\textwidth}
  \includegraphics[width=\textwidth]{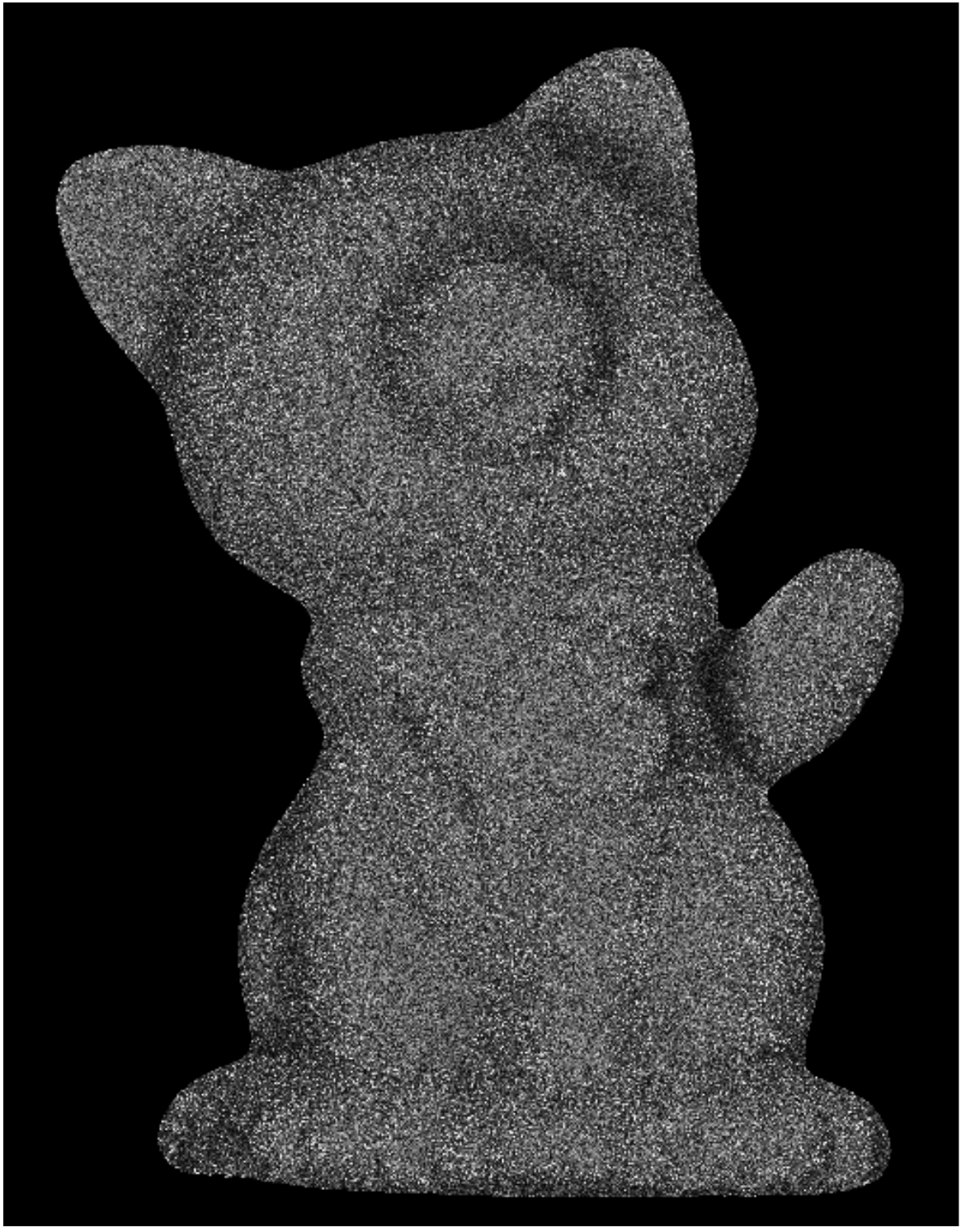}
  \caption{RPCA}
\end{subfigure}
\hspace{-2mm}
\begin{subfigure}[b]{0.105\textwidth}
  \includegraphics[width=\textwidth]{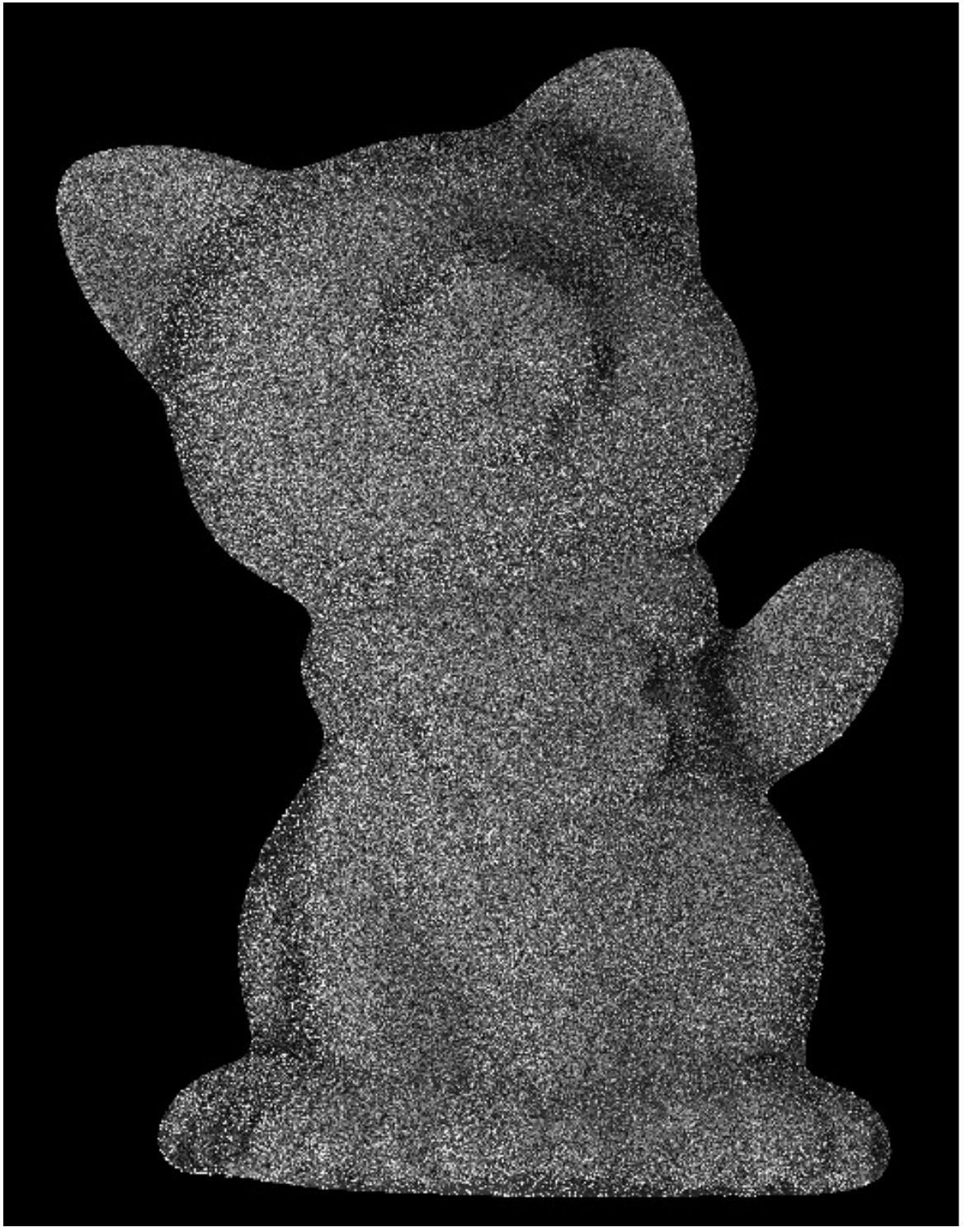}
  \caption{CBR}
\end{subfigure}
\hspace{-2mm}
\begin{subfigure}[b]{0.105\textwidth}
  \includegraphics[width=\textwidth]{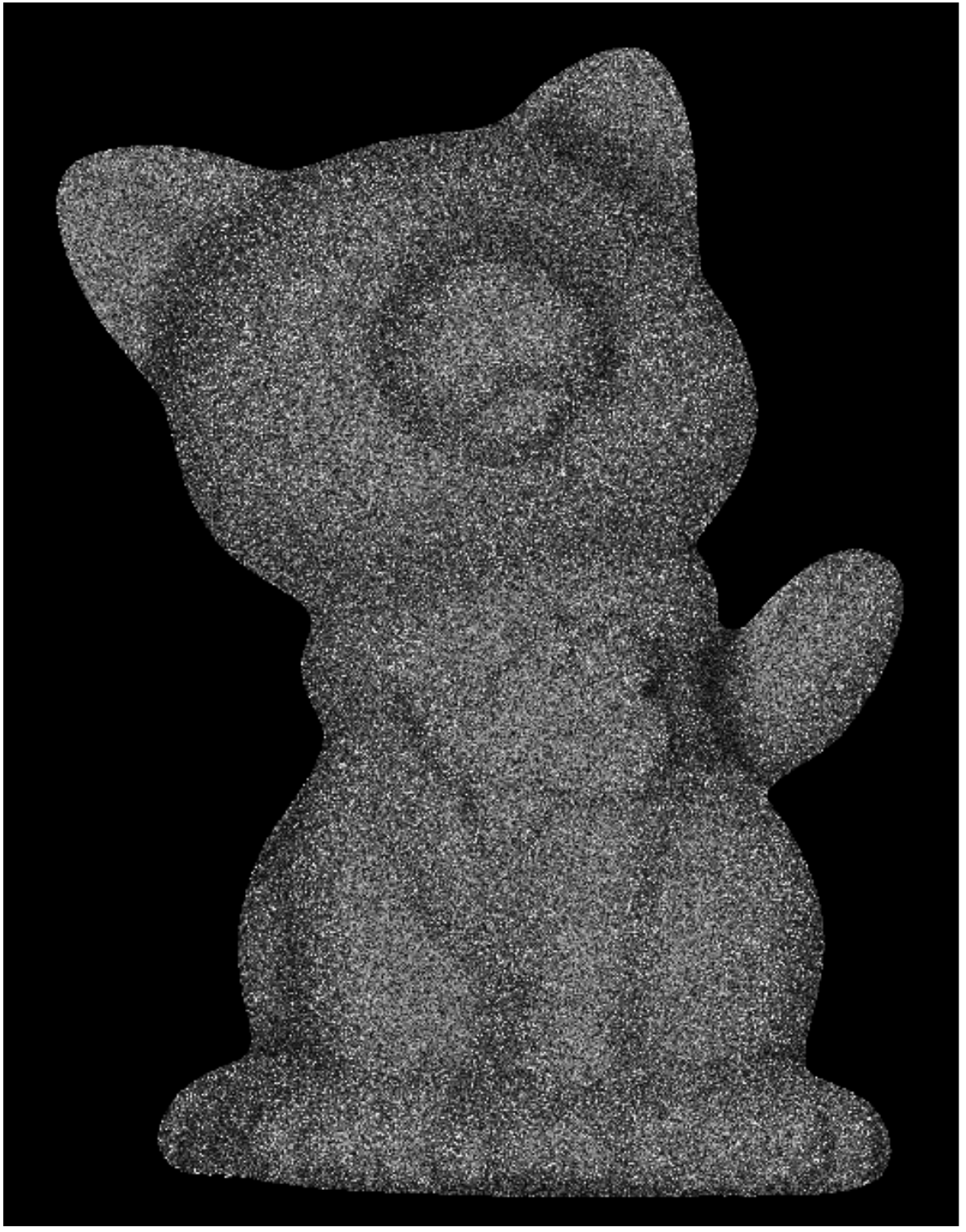}
  \caption{LS}
\end{subfigure}
\begin{subfigure}[b]{0.0275\textwidth} 
  \includegraphics[width=\textwidth]{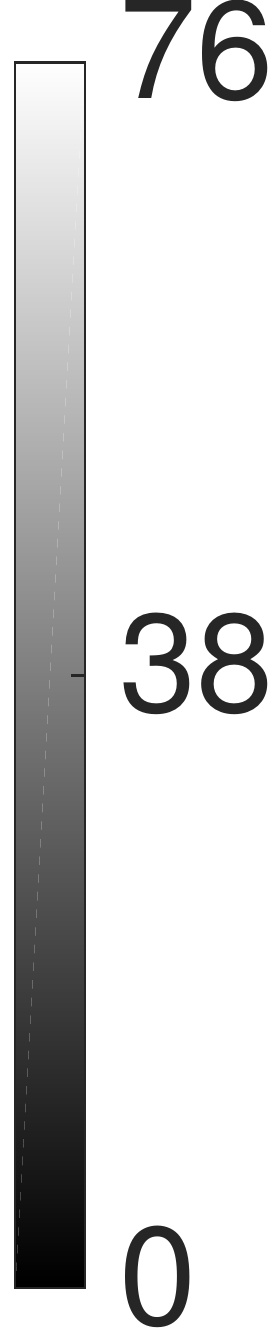}
  \caption*{}
\end{subfigure}
\caption{Normal vector plots and error maps computed from the Cat dataset \cite{xiong2015shading} with 20 images and SNR = 1 dB. Error maps plot angular error (in degrees) in the normal vectors at each point on surface.}
\label{fig:sweep_snr_20_na_0_cat_all_1_2}
\vspace{-3mm}
\end{figure*}

\subsection{Varying Noise Levels}

We first simulate the addition of Poisson noise to the images, varying the signal-to-noise-ratio (SNR). Figure~\ref{fig:sweep_snr_20_na_0_Dcats_all} illustrates the results of these simulations on a 20-image subset of the DiLiGenT Cat dataset.

As Figure~\ref{fig:sweep_snr_20_na_0_Dcats_all} shows, in the low SNR (high noise) regime, both dictionary learning approaches significantly outperform existing approaches and are able to produce much cleaner normal vectors.  The performance of all methods becomes comparable in the high SNR (low noise) regime, although our proposed dictionary learning based approaches are less sensitive to changes in the noise strength.


\subsection{Varying Number of Images}
Table~\ref{tab:sweep_nim_na_10_0_Dbears_all} illustrates the accuracy of the estimated normal vectors of each algorithm as a function of the number of images used in the reconstruction. We ran this experiment on the DiLiGenT Bear dataset, sweeping from 5 to 96 images and adding Poisson noise with 10 dB SNR.

Table 1 shows that our proposed DLNV and DLPI algorithms significantly outperform existing methods on small datasets, achieving nearly 15 degree improvements in mean angular error. These results imply that, while dictionary learning based approaches generally perform well in low SNR (high noise) regimes, they are particularly robust to noise on small datasets compared to existing methods.

\begin{table}[ht!]
\centering
\begin{tabular}{|c|c|c|c|c|c|c|}
\cline{1-7}
No. of & \multirow{2}{*}{\bf{DLPI}} & \multirow{2}{*}{\bf{DLNV}} & \multirow{2}{*}{SR} & \multirow{2}{*}{RPCA} & \multirow{2}{*}{CBR} & \multirow{2}{*}{LS} \\ 
Images & & & & & & \\
\cline{1-7}
5 & \textbf{20.91} & 21.43 & 34.29 & 33.12 & 31.10 & 34.25  \\
\cline{1-7}
15 & \textbf{9.73} & 10.20 & 14.89 & 14.44 & 14.86 & 14.90 \\
\cline{1-7}
25 & \textbf{9.23} & 9.52 & 12.12 & 11.78 & 12.76 & 12.14 \\
\cline{1-7}
35 & \textbf{9.13} & 9.18 & 11.23 & 10.93 & 11.99 & 11.24  \\
\cline{1-7}
45 & 8.96 & \textbf{8.90} & 10.62 & 10.34 & 11.71 & 10.64  \\
\cline{1-7}
55 & 8.86 & \textbf{8.78} & 10.24 & 9.97 & 11.51 & 10.25  \\
\cline{1-7}
65 & 8.89 & \textbf{8.77} & 10.00 & 9.75 & 11.29 & 10.02  \\
\cline{1-7}
75 & 8.79 & \textbf{8.68} & 9.73 & 9.50 & 11.21 & 9.750 \\
\cline{1-7}
85 & 8.72 & \textbf{8.62} & 9.57 & 9.34 & 11.20 & 9.58  \\
\cline{1-7}
96 & 8.69 & \textbf{8.61} & 9.43 & 9.22 & 11.05 & 9.45 \\
\cline{1-7}
\end{tabular}
\caption{Mean angular error sweeping number of images on the DiLiGenT Bear dataset \cite{shi2016} with SNR = 10 dB.}
\label{tab:sweep_nim_na_10_0_Dbears_all}
\vspace{-4mm}
\end{table}

\subsection{Analysis of non-DiLiGenT datasets}
In addition to the DiLiGenT dataset, we also consider the dataset from\footnote{The data can be found at \url{http://vision.seas.harvard.edu/qsfs/Data.html}} \cite{xiong2015shading}. This dataset contains images of several real objects without ground truth normal vectors. To obtain an estimate of the ground truth normal vectors, we assume the objects in this dataset follow a truly Lambertian model and compute normal vectors from the uncorrupted dataset using the simple least squares model \eqref{eq:ls}. While the Lambertian assumption may not hold exactly, the objects are matte in appearance -- the primary characteristic of Lambertian surfaces -- so these vectors are a reasonable approximation of the true normal vectors. This approach allows us to isolate the robustness of each method to noise when our data otherwise perfectly follow the modeling assumptions. Figure~\ref{fig:sweep_snr_20_na_0_hippo_all} depicts the results of these experiments.

From Figure~\ref{fig:sweep_snr_20_na_0_hippo_all}, we see that, for high SNR cases where corruptions are minimal, all methods converge to (nearly) zero mean angular error, as expected, since most methods are based on a Lambertian model. However, in the low SNR (high noise) regime, we see that, as in our previous results, both proposed dictionary learning based methods are significantly more robust to noise and produce much more accurate reconstructions. Figure~\ref{fig:sweep_snr_20_na_0_cat_all_1_2} illustrates the normal vectors obtained by the dictionary learning based approaches and error maps of all methods on the Cat dataset with an SNR of 1dB. Intuitively, the proposed adaptive dictionary learning methods are able to learn local features of data that effectively denoise the images (DLPI) or normal vectors (DLNV).

%% file: PSviaDL.bbl
\begin{thebibliography}{10}

\bibitem{woodham1980}
R.~J. Woodham,
\newblock ``Photometric method for determining surface orientation from
  multiple images,''
\newblock {\em Optical Engineering}, vol. 19, no. 1, pp. 191139--191139, 1980.

\bibitem{hayakawa1994}
H.~Hayakawa,
\newblock ``Photometric stereo under a light source with arbitrary motion,''
\newblock {\em JOSA A}, vol. 11, no. 11, pp. 3079--3089, 1994.

\bibitem{belhumeur1999}
P.~N. Belhumeur, D.~J. Kriegman, and A.~L. Yuille,
\newblock ``The bas-relief ambiguity,''
\newblock {\em International Journal of Computer Vision}, vol. 35, no. 1, pp.
  33--44, Nov. 1999.

\bibitem{yuille1999}
A.~L. Yuille, D.~Snow, R.~Epstein, and P.~N. Belhumeur,
\newblock ``Determining generative models of objects under varying
  illumination: Shape and albedo from multiple images using {SVD} and
  integrability,''
\newblock {\em International Journal of Computer Vision}, vol. 35, no. 3, pp.
  203--222, 1999.

\bibitem{georghiades2003}
A.~S. Georghiades,
\newblock ``Incorporating the torrance and sparrow model of reflectance in
  uncalibrated photometric stereo,''
\newblock in {\em ICCV}, 2003, vol.~2, pp. 816--823.

\bibitem{barsky2003}
S.~Barsky and M.~Petrou,
\newblock ``The 4-source photometric stereo technique for three-dimensional
  surfaces in the presence of highlights and shadows,''
\newblock {\em IEEE PAMI}, vol. 25, no. 10, pp. 1239--1252, Oct. 2003.

\bibitem{chandraker2007}
M.~Chandraker, S.~Agarwal, and D.~Kriegman,
\newblock ``Shadowcuts: Photometric stereo with shadows,''
\newblock in {\em CVPR}, June 2007, pp. 1--8.

\bibitem{verbiest2008}
F.~Verbiest and L.~Van Gool,
\newblock ``Photometric stereo with coherent outlier handling and confidence
  estimation,''
\newblock in {\em CVPR}, June 2008, pp. 1--8.

\bibitem{yu2010}
C.~Yu, Y.~Seo, and S.~W. Lee,
\newblock ``Photometric stereo from maximum feasible lambertian reflections,''
\newblock in {\em ECCV}, 2010, pp. 115--126.

\bibitem{wu2010}
T-P Wu and C-K Tang,
\newblock ``Photometric stereo via expectation maximization,''
\newblock {\em IEEE PAMI}, vol. 32, no. 3, pp. 546--560, Mar. 2010.

\bibitem{wu2011}
L.~Wu, A.~Ganesh, B.~Shi, Y.~Matsushita, Y.~Wang, and Y.~Ma,
\newblock ``Robust photometric stereo via low-rank matrix completion and
  recovery,''
\newblock {\em ACCV}, pp. 703--717, 2011.

\bibitem{ikehata2012}
S.~Ikehata, D.~Wipf, Y.~Matsushita, and K.~Aizawa,
\newblock ``Robust photometric stereo using sparse regression,''
\newblock in {\em CVPR}, 2012, pp. 318--325.

\bibitem{oren1995}
M.~Oren and S.~K. Nayar,
\newblock ``Generalization of the lambertian model and implications for machine
  vision,''
\newblock {\em International Journal of Computer Vision}, vol. 14, no. 3, pp.
  227--251, 1995.

\bibitem{hertzmann2005}
A.~Hertzmann and S.~M. Seitz,
\newblock ``Example-based photometric stereo: Shape reconstruction with
  general, varying {BRDFs},''
\newblock {\em IEEE PAMI}, vol. 27, no. 8, pp. 1254--1264, Aug. 2005.

\bibitem{alldrin2007_2}
N.~G. Alldrin and D.~J. Kriegman,
\newblock ``Toward reconstructing surfaces with arbitrary isotropic
  reflectance: A stratified photometric stereo approach,''
\newblock in {\em ICCV}, 2007, pp. 1--8.

\bibitem{chung2008}
H-S Chung and J.~Jia,
\newblock ``Efficient photometric stereo on glossy surfaces with wide specular
  lobes,''
\newblock in {\em CVPR}, June 2008, pp. 1--8.

\bibitem{alldrin2008}
N.~Alldrin, T.~Zickler, and D.~Kriegman,
\newblock ``Photometric stereo with non-parametric and spatially-varying
  reflectance,''
\newblock in {\em CVPR}, June 2008, pp. 1--8.

\bibitem{seitz2010}
D.~B. Goldman, B.~Curless, A.~Hertzmann, and S.~M. Seitz,
\newblock ``Shape and spatially-varying {BRDFs} from photometric stereo,''
\newblock {\em IEEE PAMI}, vol. 32, no. 6, pp. 1060--1071, 2010.

\bibitem{higo2010}
T.~Higo, Y.~Matsushita, and K.~Ikeuchi,
\newblock ``Consensus photometric stereo,''
\newblock in {\em CVPR}, June 2010, pp. 1157--1164.

\bibitem{shi2012}
B.~Shi, P.~Tan, Y.~Matsushita, and K.~Ikeuchi,
\newblock ``Elevation angle from reflectance monotonicity: Photometric stereo
  for general isotropic reflectances,''
\newblock {\em ECCV}, pp. 455--468, 2012.

\bibitem{chandraker2013}
M.~Chandraker, J.~Bai, and R.~Ramamoorthi,
\newblock ``On differential photometric reconstruction for unknown, isotropic
  {BRDFs},''
\newblock {\em IEEE PAMI}, vol. 35, no. 12, pp. 2941--2955, 2013.

\bibitem{ikehata2014}
S.~Ikehata and K.~Aizawa,
\newblock ``Photometric stereo using constrained bivariate regression for
  general isotropic surfaces,''
\newblock in {\em CVPR}, 2014, pp. 2179--2186.

\bibitem{shi2014}
B.~Shi, P.~Tan, Y.~Matsushita, and K.~Ikeuchi,
\newblock ``Bi-polynomial modeling of low-frequency reflectances,''
\newblock {\em IEEE PAMI}, vol. 36, no. 6, pp. 1078--1091, June 2014.

\bibitem{elad2006image}
M.~Elad and M.~Aharon,
\newblock ``Image denoising via sparse and redundant representations over
  learned dictionaries,''
\newblock {\em IEEE Trans. on Image Proc.}, vol. 15, no. 12, pp. 3736--3745,
  2006.

\bibitem{aharon2006rm}
M.~Aharon, M.~Elad, and A.~Bruckstein,
\newblock ``$k$-svd: An algorithm for designing overcomplete dictionaries for
  sparse representation,''
\newblock {\em IEEE Trans. on Signal Proc.}, vol. 54, no. 11, pp. 4311--4322,
  2006.

\bibitem{ravishankar2011mr}
S.~Ravishankar and Y.~Bresler,
\newblock ``{MR} image reconstruction from highly undersampled k-space data by
  dictionary learning,''
\newblock {\em IEEE Trans. on Med. Imag.}, vol. 30, no. 5, pp. 1028--1041,
  2011.

\bibitem{ravishankar2016lassi}
S.~Ravishankar, B.~E. Moore, R.~R. Nadakuditi, and J.~A. Fessler,
\newblock ``{LASSI}: A low-rank and adaptive sparse signal model for highly
  accelerated dynamic imaging,''
\newblock in {\em IVMSP Workshop}, 2016, pp. 1--5.

\bibitem{shi2016}
B.~Shi, Z.~Wu, Z.~Mo, D.~Duan, S-K Yeung, and P.~Tan,
\newblock ``A benchmark dataset and evaluation for non-lambertian and
  uncalibrated photometric stereo,''
\newblock in {\em CVPR}, 2016.

\bibitem{simchony1990}
T.~Simchony, R.~Chellappa, and M.~Shao,
\newblock ``Direct analytical methods for solving poisson equations in computer
  vision problems,''
\newblock {\em IEEE PAMI}, vol. 12, no. 5, pp. 435--446, May 1990.

\bibitem{sairajfes2}
S.~Ravishankar, R.~R. Nadakuditi, and J.~A. Fessler,
\newblock ``Efficient sum of outer products dictionary learning {(SOUP-DIL)} -
  the $\ell_{0}$ method,''
\newblock {\em arXiv preprint arXiv:1511.08842}, 2015.

\bibitem{kar}
R.~Gribonval and K.~Schnass,
\newblock ``Dictionary identification--sparse matrix-factorization via
  $\textit{l}_{1}$ -minimization,''
\newblock {\em IEEE Trans. on Inform. Theory}, vol. 56, no. 7, pp. 3523--3539,
  2010.

\bibitem{dinokat2016}
S.~Ravishankar, B.~E. Moore, R.~R. Nadakuditi, and J.~A. Fessler,
\newblock ``Efficient learning of dictionaries with low-rank atoms,''
\newblock in {\em Proc. IEEE Global Conference on Signal and Information
  Processing}, 2016.

\bibitem{parboyd}
N.~Parikh and S.~Boyd,
\newblock ``Proximal algorithms,''
\newblock {\em Found. Trends Optim.}, vol. 1, no. 3, pp. 127--239, Jan. 2014.

\bibitem{xiong2015shading}
Y.~Xiong, A.~Chakrabarti, R.~Basri, S.~J. Gortler, D.~W. Jacobs, and T.~E.
  Zickler,
\newblock ``From shading to local shape,''
\newblock {\em IEEE PAMI}, vol. 37, no. 1, pp. 67--79, 2015.

\end{thebibliography}
